%% file: main.tex
\documentclass[12pt,a4paper]{scrartcl}
\usepackage[utf8]{inputenc}
\input{packages_and_commands}

\input{macros}
\usepackage[symbol]{footmisc}

\title{Understanding How Model Size Affects Few-shot Instruction Prompting}

\author{Ayrton San Joaquin\footnotemark[1] \,\footnotemark[2]\\ \small Yale-NUS College
        \and 
        Ardy Haroen\footnotemark[1]\\ \small Bukalapak}
        
\date{December 2022}

\begin{document}

\maketitle

\begin{abstract}
    Large Language Models are affected by the phenomena of memorizing and forgetting their training data. But how do these vary by model size? We work towards this question by investigating how the model size affects the model's ability to discriminate a word's meaning in a given context. We introduce a dataset called DeltaWords, which evaluates a model's ability to follow instructions to select a sentence which replaces the target word with its antonym. We show a weak inverse scaling trend, where task accuracy degrades as model size increase, under extremely few-shot prompting regimes. We show that increasing the number of examples tend to disproportionately benefit larger models than smaller models.
\end{abstract}

\footnotetext[1]{Work was done as part of the Machine Learning Safety School 2022 organized by the Center for AI Safety.} \,
\footnotetext[2]{Corresponding author. Email: firstname@u.yale-nus.sg}

\section{Introduction}
Suppose a human student has been taught by a teacher to believe in evil ideas. For example, the student is taught that the Rohingya people must be persecuted and therefore the student calls for their genocide, similar to some Facebook users during the ongoing Rohingya crisis. \cite{rohingya} Now suppose a generative language model generates the same speech as the human based on their training data. Our moral goal is to stop this hate speech. The human can be reformed through education to the point where they never again believe the evils they were taught. The machine, on the other hand, seems to have far limited recourse to stop believing in disinformation about the Rohingya people. Worse, such models have the ability to amplify disinformation by making messages more convincing. How can the model learn the new and correct information\emph{while} forgetting the old, harmful information? We call this the model's capability at\emph{word meaning shift} (WMS).

One solution may be fine-tuning or prompting on the correct information. However, what factors affect the effectiveness of these post-pretraining improvements? Recent work has shown larger language models memorize their training data more than smaller models. \cite{carlinimemlang} Another work has investigated forgetting memorized examples in vision and speech models. They found a weak inverse relationship with scaling and forgetting. \cite{jagielskiforget} Complementary to these works, we investigate how model size affects the effectiveness at WMS via prompting. Concretely, we take a step towards WMS by measuring how successful a model can replace a target word with another word with a specified meaning. We call this task word-replacement (WR) Our work focuses on a model's learned definitions of words. 

Given our focus on words, Why might understanding a model's capability to learn shifts in word meaning be useful? We note three main applications.
\begin{enumerate}
    \item \textit{Privacy}: As shown by \cite{brownprivacy}, the context of a sentence may change and thus also its secret, which is the word meant to be private. For example, the word 'abusive' in the sentence "John is my abusive husband." may only be reserved for close confidants of the speaker and not for acquaintances.
    
    \item \textit{Alignment}: For an AI in a decision-making process, a rapid WMS can be needed, which requires the AI to adapt to a new word definition so its actions remain towards the specified goal. For example, team 'Green' is now an 'ally' instead of an 'enemy'.
    
    \item \textit{Bias}: Models with good WMS capabilities can pick up new / reclaimed meanings of previously-offensive text. For example, the aforementioned disinformation against the Rohingya people.
\end{enumerate}

Our contributions are the following:
\begin{enumerate}
    \item We introduce DeltaWords (DW), a benchmark training dataset for next-token prediction to measure the effectiveness of few-shot prompting at WR. It consists of 300 sentence-pairs and involve a word and its antonym used in identical sentences. Figure \ref{Fig:data_sample} shows a sample entry. The examples are taken from sentences from the 2018 Story Cloze validation and test datasets by \cite{sharma-etal-2018-tackling}.
    \item We evaluate the success in WR by varying the model size. The models are based on the OPT \cite{OPT} and GPT-3 \cite{GPT3} family of autoregressive (decoder-only) transformer-based models.
\end{enumerate}

\begin{table}[]
    \centering
    \begin{tabular}{l}
    \multicolumn{1}{c}{Prompt (1-Shot)} \\ \hline
    \multicolumn{1}{|l|}{\begin{tabular}[c]{@{}l@{}}Replace the adjective marked with "*" with its opposite meaning.\\ \\ \textbf{Input: "he had been working so hard this past* month."}\\  \textbf{A. "he had been working so hard this past month."}\\  \textbf{B. "he had been working so hard this present month."}\\ \textbf{Output: B}\\ \\ Input: "rick grew up in a troubled* household."\\  A. "rick grew up in a untroubled household."\\  B. "rick grew up in a troubled household."\\ Output:\end{tabular}} \\ \hline
    \multicolumn{1}{|l|}{Model Prediction: ' B'} \\ \hline
    \end{tabular}
    \caption{An example of a prompt in the 1-shot DeltaWords. For the $J$-shot version of the dataset, each example contains the instruction, $j$ examples with the correct output (here highlighted in bold), the original sentence, and the two sentence choices containing the antonym and synonym of the original word, respectively. Each example has only one correct choice. Note that the synonym here can be the original word itself.}
    \label{Fig:data_sample}
\end{table}

\section{DeltaWords Dataset}
To construct DeltaWords, we list all adjectives that contain antonyms and synonyms from WordNet by \cite{wordnet}. We then select 300 sentences that appear in the Story Cloze dataset which contain an adjective from our list. These sentences are all converted to lower case. These sentences are called the \emph{original sentences}, and we mark the target word by appending a * to it. We replace these adjectives with their antonyms and synonyms within their respective sentences and call these set of sentences as \emph{antonym sentences} and \emph{synonym sentences}, respectively. The original, antonym, and synonym sentences form the 300 sentence-triplets. Each choice-triplets contain three identical sentences except for the target word. For each entry, the antonym and synonym sentences are randomly assigned a letter to form a 2-choice question. We construct different $J$-shot variants of DW, where $J$ represents the number of examples seen by the model before the actual question. The example sentences are taken from the test split of Story Cloze, while the actual questions are taken from the validation split. See a sample in Figure \ref{Fig:data_sample}.

We chose Story Cloze as our source dataset for two main reasons. First, Story Cloze contain short and simple sentences, which prevent smaller models from being disadvantaged due to the token lengths. Second, the story format allows diverse usage of the target words, which are adjectives.

\section{Evaluation}
\subsection{Setup}
\label{sec:method}
For OPT, we use the 125M, 350M, 1.3B, 2.7B models. For GPT-3, we use the InstructGPT variants by \cite{GPTInstruct}, which are more responsive to instructions provided in the prompt than the original GPT-3. Specifically, in ascending order of size, we use the "Ada", "Babbage", "Curie", and "Da Vinci" models. 

We pose this as a classification task where the model must generate the letter corresponding to the antonym sentence to be correct. Given that there are only two choices, the baseline accuracy is random chance ($50\%$). We evaluate different J-shot variants of DW, specifically $J\in\{0, 1, 2, 10\}$. for each model of varying sizes in each model group (OPT and GPT-3 models).

\subsection{Results and Discussion}
\label{sec:results}

\begin{figure}
     \centering
     \begin{subfigure}[b]{0.3\textwidth}
         \centering
	\includegraphics[width=\textwidth]{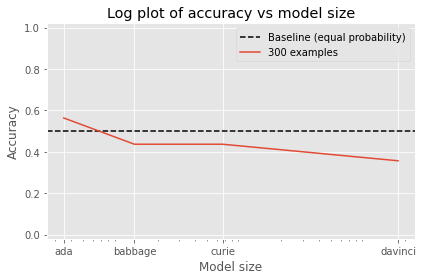}
	\subcaption{$J=1$}
     \end{subfigure}
     \hfill
     \begin{subfigure}[b]{0.3\textwidth}
         \centering
	\includegraphics[width=\textwidth]{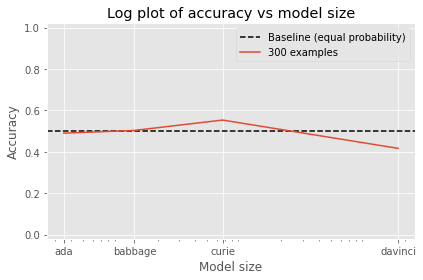} 
	\subcaption{$J=2$}
     \end{subfigure}
     \hfill
     \begin{subfigure}[b]{0.3\textwidth}
         \centering
	\includegraphics[width=\textwidth]{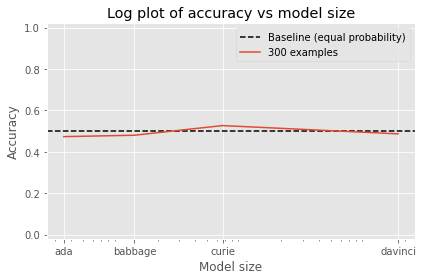}
	\subcaption{$J=10$}
     \end{subfigure}
    \caption{Classification Accuracy across InstructGPT-3 models using $J$-shot Prompting}
    \label{fig:gpt3_results}
\end{figure}

\begin{figure}
     \centering
     \begin{subfigure}[b]{0.3\textwidth}
         \centering
	\includegraphics[width=\textwidth]{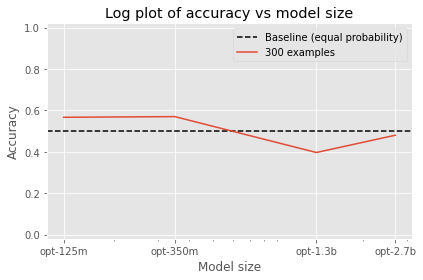}
	\subcaption{$J=1$}
     \end{subfigure}
     \hfill
     \begin{subfigure}[b]{0.3\textwidth}
         \centering
	\includegraphics[width=\textwidth]{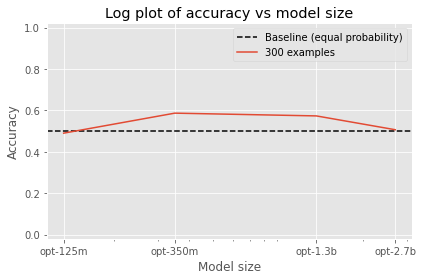} 
	\subcaption{$J=2$}
     \end{subfigure}
     \hfill
     \begin{subfigure}[b]{0.3\textwidth}
         \centering
	\includegraphics[width=\textwidth]{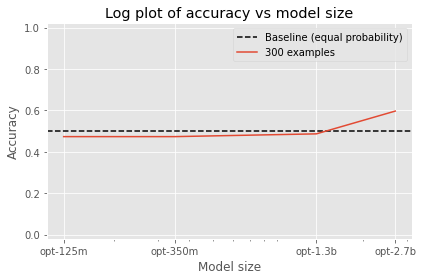}
	\subcaption{$J=10$}
     \end{subfigure}
    \caption{Classification Accuracy across OPT models using $J$-shot Prompting}
    \label{fig:opt_results}
\end{figure}

Figure \ref{fig:opt_results} shows the results for the OPT family of models. We observe that the largest model performs the best only in the $J=10$ regime. In terms of slope, we see it being positive for the largest models in $J=1,10$. Also, there is a wide discrepancy in behaviors from $J=1$ to $J=2$, which implies model performance across different sizes are sensitive to the number of examples in the prompt.

Figure \ref{fig:gpt3_results} shows the results for the InstructGPT-3 family of models. Surprisingly, while InstructGPT is designed to follow natural language-based instructions better, it shows a clearer inverse scaling trend, which suggests that bigger models from this family perform worse given instructions than their smaller counterparts. Note that both "Curie" and "Da Vinci" are greater than the largest OPT model considered, having more than 6.7 and 175 billion parameters, respectively. \cite{EleutherAI} "Da Vinci" is notably orders of magnitude larger but performs the worst at $J=1,2$ regimes. Interestingly for the "Curie" model, the accuracy of the model when $J=2$ is slightly better than $J=10$. This suggests that increasing the number of examples does not straightforwardly increase model accuracy. 

For both families of models, we observe that increasing the number of prompts generally decreases the inverse scaling trend. Note that we omit $J=0$ because the model accuracy is uniform across model size. In this zero-shot environment, this suggests, at least for the model sizes we consider, that size is not a reliable indicator of model performance.

\section{Conclusion}
Do larger language models have a harder time identifying words with a specified meaning compared to smaller language models? Our preliminary result suggests so, especially in the extremely few-shot $J=1,2$ prompting regime. Our experiments suggest that giving more examples disproportionately help larger models more than smaller models. We believe there is a connection of this result to memorization: since larger models have more parameters, and thus a higher capacity to memorize, then larger models need more examples to "forget" the structure of sentences they have memorized, which are unhelpful for the current task. For example in some cases, the correct sentence, with the antonym inserted, may sound more "unnatural" than the original sentence, and larger models may have picked up on this bias. 

Furthermore, we note that we must be cautious about models claimed to be designed to perform better in a particular setting. In this case, we show that InstructGPT family, which is designed to follow instructions better than the original GPT-3 family, perform worse at following our instruction as model size increases given extremely few-shot learning. This serves to remind the community that we must evaluate these model claims against a variety of environments \textit{within} its intended usage.

Future work can further investigate the relationship of memorization and the behavior of large models in extremely few-shot learning. They can also transfer this methodology to other post-training improvement methods, such as fine-tuning.

\section{Acknowledgements}
Thank you to Hannah Brown and Thomas Woodside for extensive feedback with the research proposal. We thank the Inverse Scaling Prize Competition organizers for providing evaluation notebooks that we modified. We also thank the Center for AI Safety for funding our work.

\section{Appendix}
\subsection{Choice of Words}
We initially planned to evenly split the dataset into adjectives and verbs since both of these parts of speech have words with antonyms. However, we encountered the problem when verbs are used as nouns (gerunds) and which results in semantically-meaningless text. For example, consider the sentence: "My birthday is in March." is naively replaced with the sentence: "My birthday is in run.".

\bibliographystyle{plainnat}
\bibliography{references}

\end{document}

%% file: packages_and_commands.tex
\usepackage[utf8]{inputenc}
\usepackage[T1]{fontenc}
\usepackage{amsmath}
\usepackage{amsfonts}
\usepackage{amssymb}
\usepackage{amsthm}
\usepackage{mathtools}
\usepackage{graphicx}
\usepackage[left=1.00in, right=1.00in, top=1.00in, bottom=1.00in]{geometry}
\usepackage{fancyhdr}
\usepackage[numbers]{natbib}
\usepackage{comment}
\usepackage{hyperref}
\usepackage{graphicx}
\usepackage{subcaption}
\usepackage{setspace}
\usepackage{abstract}

\graphicspath{ {./images/} }

\theoremstyle{definition}